\makeatletter
\@namedef{ver@everyshi.sty}{}
\makeatother

\def\cvprPaperID{0000}
\def\confName{CVPR}
\def\confYear{2023}

\def\paperTitle{LDFA: Latent Diffusion Face Anonymization for Self-driving Applications}

\def\authorBlock{
    Marvin Klemp$^1$ \qquad
    Kevin R\"osch$^{1,2}$ \qquad
    Royden Wagner$^1$  \qquad
    Jannik Quehl$^1$ \qquad
    Martin Lauer$^1$\\
    $^1$Karlsruhe Institute of Technology \qquad $^2$ FZI Research Center for Information Technology\\
    {\tt\small \{firstname.lastname\}@kit.edu}
}

\newif\ifreview 
\newif\ifarxiv \newcommand{\arxiv}{\arxivtrue}
\newif\ifcamera 
\newif\ifrebuttal 

\arxiv
\pdfoutput=1
\documentclass[10pt,twocolumn,letterpaper]{article}
\ifreview \usepackage[review]{cvpr} \fi
\ifarxiv \usepackage[pagenumbers]{cvpr} \fi
\ifrebuttal \usepackage[rebuttal]{cvpr} \fi
\ifcamera \usepackage{cvpr} \fi

\usepackage{graphicx}
\usepackage{amsmath}
\usepackage{amssymb}
\usepackage{booktabs}
\usepackage{todonotes}

\makeatletter
\@namedef{ver@everyshi.sty}{}
\makeatother
\usepackage{tikz}

\usepackage{times}
\usepackage{microtype}
\usepackage{epsfig}
\usepackage{caption}
\usepackage{svg}
\usepackage{float}
\usepackage{placeins}
\usepackage{color, colortbl}
\usepackage{stfloats}
\usepackage{enumitem}
\usepackage{tabularx}
\usepackage{xstring}
\usepackage{multirow}
\usepackage{xspace}
\usepackage{url}
\usepackage{subcaption}
\usepackage[hang,flushmargin]{footmisc}

\ifcamera \usepackage[accsupp]{axessibility} \fi

% ...

\ifarxiv  \fi

\newcommand{\R}[1]{{%
    \textbf{%
        \ifstrequal{#1}{1}{\textcolor{red}{R#1}}{%
        \ifstrequal{#1}{2}{\textcolor{blue}{R#1}}{%
        \ifstrequal{#1}{3}{\textcolor{magenta}{R#1}}{%
        \ifstrequal{#1}{4}{\textcolor{teal}{R#1}}{%
                           \textcolor{cyan}{R#1}%
        }}}}%
    }%
}}

\usepackage{xr-hyper}

\makeatletter
\newcommand*{\addFileDependency}[1]{
  \typeout{(#1)}
  \@addtofilelist{#1}
  \IfFileExists{#1}{}{\typeout{No file #1.}}
}

\makeatother

\usepackage[pagebackref,breaklinks,colorlinks]{hyperref}
\usepackage[capitalize]{cleveref}
\crefname{section}{Sec.}{Secs.}
\crefname{table}{Table}{Tables}
\crefname{figure}{Fig.}{Figs.}

\begin{document}
\title{\paperTitle}
\author{\authorBlock}
\twocolumn[{%
\renewcommand\twocolumn[1][]{#1}%
\maketitle
\begin{center}
    \centering
    \captionsetup{type=figure}
    \includegraphics[width=\linewidth]{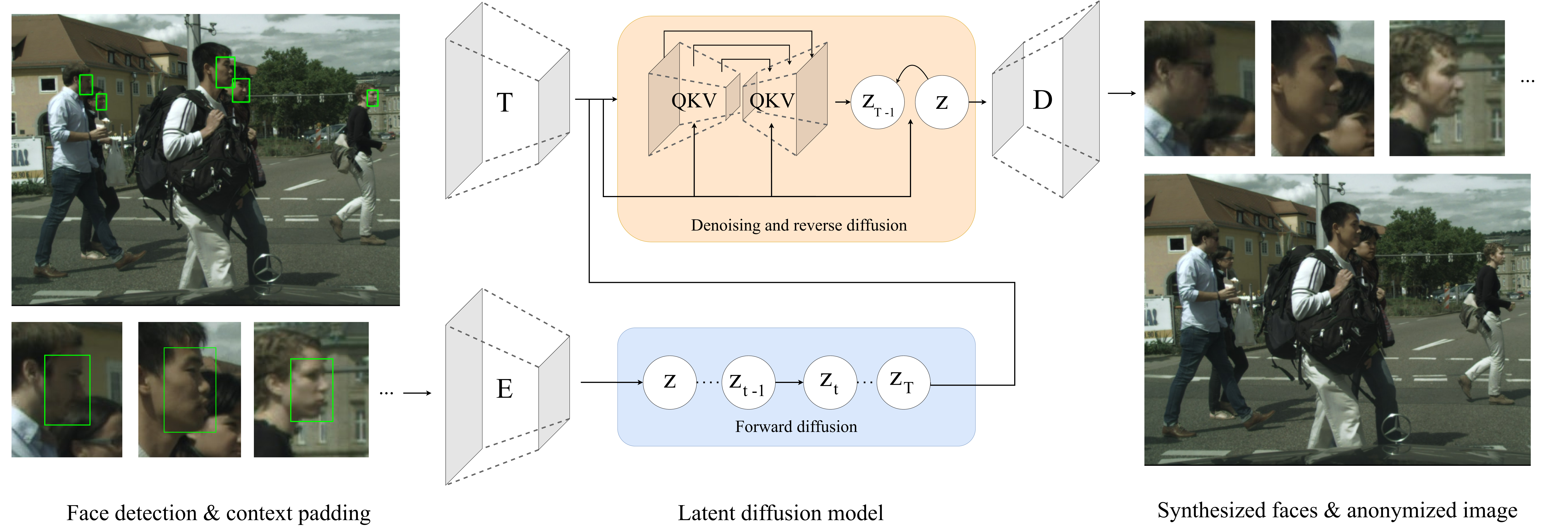}
    \captionof{figure}{\textbf{LDFA pipeline.} Faces are detected by a RetinaFace detector \cite{retinaface}, afterwards, detections are padded to provide a latent diffusion model \cite{rombach2022high} with context to synthesize realistic faces. Finally, the synthesized faces are inserted in the input image to generate an anomymized image.
Further details are in Section \ref{sec:method}.}
    \label{fig:ldfa-pipeline}
\end{center}%
}]


\begin{abstract}
In order to protect vulnerable road users (VRUs), such as pedestrians or cyclists, it is essential that intelligent transportation systems (ITS) accurately identify them.
Therefore, datasets used to train perception models of ITS must contain a significant number of vulnerable road users.
However, data protection regulations require that individuals are anonymized in such datasets.
In this work, we introduce a novel deep learning-based pipeline for face anonymization in the context of ITS.
In contrast to related methods, we do not use generative adversarial networks (GANs) but build upon recent advances in diffusion models.
We propose a two-stage method, which contains a face detection model followed by a latent diffusion model to generate realistic face in-paintings.
To demonstrate the versatility of anonymized images, we train segmentation methods on anonymized data and evaluate them on non-anonymized data. 
Our experiment reveal that our pipeline is better suited to anonymize data for segmentation than naive methods and performes comparably with recent GAN-based methods.
Moreover, face detectors achieve higher mAP scores for faces anonymized by our method compared to naive or recent GAN-based methods.
\end{abstract}
\section{Introduction}
\label{sec:intro}
In the context of intelligent transportation systems (ITS), pedestrians and cyclists are classified as vulnerable road users \cite{mannion2019vulnerable}. 
In case of an accident, they have little to no protection from the impact forces.
Therefore, it is crucial that an ITS accurately and robustly detects such vulnerable road users (VRUs) in order to protect them.
Hence, datasets used to train perception models of ITS are required to contain a significant number of pedestrians and cyclists.
But VRUs should not only be protected physically, as of 25th May 2018, the general data protection regulation (GDPR) came to effect in Europe \cite{gdpr-eu} to also protect their data privacy.
The GDPR affects all processing of personal data including images of individuals and requires the consent of the individual for processing their data.
This complicates the creation of datasets focused on perception in urban scenes, such as Cityscapes \cite{cordts2016cityscapes} or Mapillary Vistas \cite{neuhold2017mapillary}.
Since obtaining the consent of all individuals from whom such a dataset contains recordings is impractical, the recordings must be anonymized.
However, naive face anonymization techniques, such as blurring or cropping, can reduce the performance of deep learning models for perception \cite{zhou2022impacts}. 
For instance, perception models trained on datasets with blurring as face anonymization method learn a representation of humans with blurred faces.
This limits their ability to generalize to real world scenarios without anonymization. \cite{klomp2021safefakes}
Consequently, anonymization techniques, which produce images that are as realistic as possible are superior \cite{hukkelaas2019deepprivacy}.
For this purpose, deep learning methods are used that replace faces by artificially generated faces \cite{hukkelaas2019deepprivacy, maximov2020ciagan, hukkelaas2023deepprivacy2}.
These methods require complex semantic reasoning to recognize faces and their poses and subsequently replace them with similar artificially generated faces.

In this work, we introduce a novel deep learning-based pipeline for face anonymization in the context of ITS.
In contrast to related methods, we do not use generative adversarial networks (GANs), but build upon recent advances in diffusion models \cite{rombach2022high}. 
Our main contributions can be summarized as follows:
\begin{itemize}
    \item We introduce a two-stage pipeline using a face detection model followed by a latent diffusion model to generate realistic face anonymization.
    \item We show that general diffusion models are equally fitted for face anonymization in comparison to recent, specialized face anonymization methods.
    \item We show that the mAP of a face detection network inferred on images, which are anonymized with LDFA, is higher compared to recent GAN-based methods.
\end{itemize}

\section{Related Work}
\label{sec:related}
\textbf{Anonymization for ITS.}
Zhou and Beyerer \cite{zhou2022impacts} train semantic segmentation models on anonymized versions of the Cityscapes dataset.
Their experiments reveal that face anonymization using blurring or cropping degrades segmentation performance, while face in-painting using GANs does not effect the segmentation performance. 
Geyer et al. \cite{geyer2020a2d2} and Wilson et al. \cite{wilson2023argoverse} use naive anonymization techniques such as blurring or pixelization to provide privacy-preserving datasets for ITS development.
These techniques are well suited to anonymize license plates \cite{schnabel2019impact} but degrade the segmentation performance when used to anonymize faces \cite{zhou2022impacts}.

\textbf{Recent GAN models for face anonymization.}
DeepPrivacy \cite{hukkelaas2019deepprivacy} and DeepPrivacy2 \cite{hukkelaas2023deepprivacy2} are closely related methods that jointly perform face detection, artifical face generation, and face in-painting.
DeepPrivacy detects faces using DSDF \cite{li2019dsfd} and estimates face keypoints using a modified Mask R-CNN \cite{he2017mask}.
Artificial faces are generated by a conditional GAN model \cite{isola2017image} conditioned on the surroundings of faces and estimated face keypoints.
DeepPrivacy2 detects and segments faces or whole bodies using an ensemble of three detectors, DSDF, CSE \cite{neverova2020continuous}, and Mask-RCNN.
Artificial faces are generated by a style-based generator inspired by StyleGAN2 \cite{karras2020analyzing}.

CIAGAN \cite{maximov2020ciagan} detects face keypoints using a histogram of oriented gradients (HOG) model. 
Face keypoints are further processed to generate an abstract face landmark image, which contains pose and expression.
Face landmark images and face surroundings are used to guide a conditional GAN for face generation.

These GAN-based methods need to be trained or fine-tuned on face-specific datasets (e.g., CelebA \cite{liu2015deep}) to generate realistic artifical faces.
In contrast, recent general pre-trained diffusion models \cite{rombach2022high, lugmayr2022repaint} can generate realistic face in-paintings without further fine-tuning.

\section{Method}
\label{sec:method}
\subsection{Diffusion Models}
Diffusion models (DMs) are generative models that recently showed particularly great success in the domain of image generation \cite{ramesh2021zero, ramesh2022hierarchical, saharia2022photorealistic}. 
But because of how adaptable they are, they can also be used in other fields, like point cloud generation \cite{luo2021diffusion} or audio generation \cite{kong2020diffwave}. 
During training, a forward diffusion process based on a Markov Chain is used to incrementally add noise to the input data.
Afterwards, a neural network is trained to reverse this process and reconstruct data from noisy representations $z_T$, see Figure \ref{fig:ldfa-pipeline}.
During inference, input data can be transformed into a noisy latent representation in the same way.
The amount of added noise controls how much information from the original data is used to synthesize new data.
Alternatively, random noise can be used as input and a transformer $T$ to encode multimodal data (e.g., text, semantic maps, images) for conditioning.
Conditioning signals are fed into the denoising U-Net using a cross-attention mechanism ($QKV$) to guide the reverse diffusion process.

\subsection{Latent Diffusion Models}
In contrast to regular DMs, which work directly on the pixel space, latent diffusion models (LDMs) \cite{rombach2022high} first transform the input image into a latent feature space and then perform the diffusion process within this lower dimensional space.
This is realized by training an autoencoder, which reduces the dimensionality of the input data considerably to a more efficient representation in latent space.
The autoencoder consists of an encoder $E$ and a decoder $D$ (cf. Figure \ref{fig:ldfa-pipeline}).
This allows for better scaling properties with respect to the spatial dimensionality as well as for more efficient training and evaluation of the resulting model \cite{rombach2022high}.

\subsection{Face Anonymization using Latent Diffusion}
Our pipeline starts by detecting the faces which are supposed to be anonymized. 
For this, we use \textsc{RetinaFace}, which achieves state-of-the-art results for face detection on the WIDER FACE dataset \cite{retinaface, serengil2021lightface}. 
As DMs are able to generate arbitrary patches of an image, we favor high recall over high precision in detecting faces and use a low detection confidence threshold of 0.4.
Once faces are detected, one image-to-image process is started for each detected bounding box. 
Each face is padded by 32 pixels before it is cut, see Figure \ref{fig:ldfa-pipeline}.
The padded area provides context for the scene during the reverse diffusion process.
However, only the facial region without padding is used to replace the contents of the original image. 
We re-scale the image patch to $512^2$ pixels and pass the resized image patch into the image-to-image LDM.
The full pipeline of our method is visualized in Figure \ref{fig:ldfa-pipeline}.
We parameterize the LDM as follows: we use the \textit{stable-diffusion-2-inpainting} weights \cite{rombach2022high}, no prompt, a CFG-Scale of 1, the \textit{k\_euler\_a} sampler and 50 inference steps. 
This process is repeated for each face individually. 

\begin{figure}
\centering
    \includegraphics[width=0.95\linewidth]{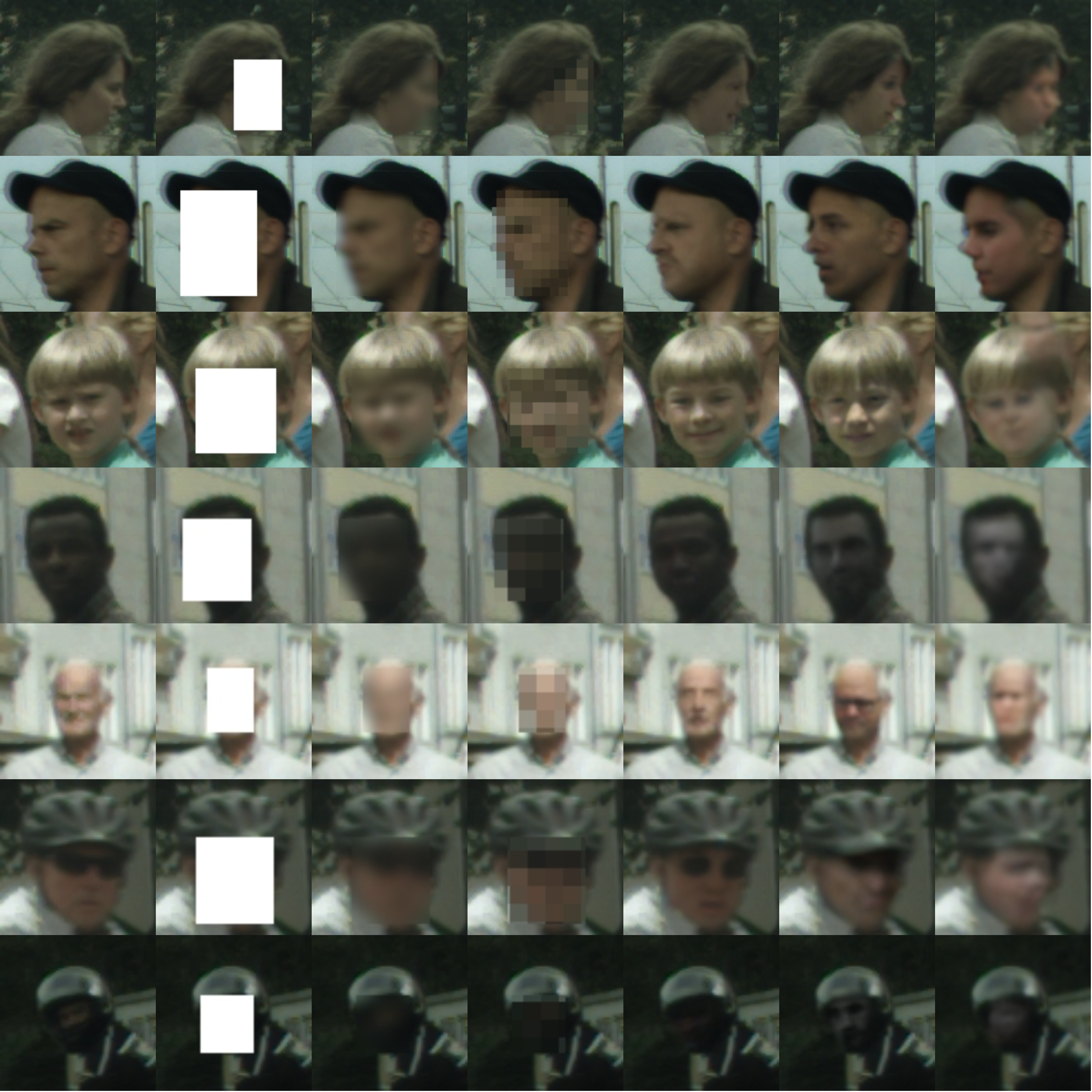}
    \caption{\textbf{Application of anonymization methods in different situations}. The applied methods (from left to right) are: None (original image), \textsc{Crop}, \textsc{Gauss}, \textsc{Pixel}, \textsc{LDFA}, \textsc{DeepPrivacy1}, \textsc{DeepPrivacy2}. We advise the reader to zoom in.}
    \label{fig:matrix}
\end{figure}

\section{Experiments}
\subsection{Anonymization Methods}
In the following experiments, we compare various anonymization methods. 
We consider deep learning based methods \textsc{DeepPrivacy}, and \textsc{DeepPrivacy2} which are state-of-the-art methods for face and full-body anonymization. 
\textsc{DeepPrivacy} is used in the default, and \textsc{DeepPrivacy2} in the face-anonymization configuration.
Additionally, we evaluate our LDM-based pipeline (\textsc{LDFA}).
Next to deep learning based methods, we compare naive machine vision based anonymization methods.
The \textsc{Gauss} method applies a Gaussian filter ($\mu = 0, \sigma = 3$).
\textsc{Crop} entirely crops the facial contents and replaces them by the maximal value for all channels.
The \textsc{Pixel} method performs pixelization as anonymization. 
In detail, face areas are split into 8x8 pixel patches and each patch is assigend the mean value of this patch.
Figure \ref{fig:matrix} shows faces, synthesised by our method, naive, and comparable state-of-the-art methods in a variety of different scenarios which may occur in an autonomous dataset.
For our experiments we used the Cityscapes dataset as a showcase \cite{cordts2016cityscapes}.

\subsection{Training Segmentation Models on Anonymized Images}
In this experiment, we evaluate the impact of anonymization methods on semantic segmentation models.
Therefore, we train several segmentation models on differently anonymized training sets and evaluate them on the same non-anonymized validation set.
As dataset, we use the Cityscapes dataset, which contains recordings of urban driving in European cities.
We use the official training split for training, and the validation split is used as a test set.

\textbf{Experimental setup.} As segmentation model, we use the \textsc{Mask2Former} model \cite{cheng2022masked}.
\textsc{Mask2Former} models are state-of-the-art segmentation models based on the vision transformer (ViT) \cite{dosovitskiyimage} and detection transformer (DETR) \cite{carion2020end} architectures.
We use the semantic segmentation configuration with a ResNet-50 \cite{he2016deep} backbone.
All models are trained for 54 epochs with a batch size of 16.
We chose \textsc{AdamW} \cite{loshchilovdecoupled} as optimizer with an initial learning rate of $1^{-5}$.
In addition, we compare our results with the results from a fully-convolutional model \textsc{PSPNet} \cite{zhao2017pyramid} reported by Zhou and Beyerer \cite{zhou2022impacts}.

\textbf{Evaluation metrics.} To evaluate the segmentation performance, we are using the intersection-over-union (IoU), the realtive IoU change w.r.t baseline models ($\Delta\text{IoU}^{\text{rel}}$), and the instance-level IoU (iIoU) metric.
The $\Delta\text{IoU}^{\text{rel}}$ is computed as
\begin{equation}
    \Delta\text{IoU}^{\text{rel}} = \frac{\text{IoU}_\text{anon} - \text{IoU}_\text{base}}{\text{IoU}_\text{base}},
\end{equation}
where $\text{IoU}_\text{anon}$ is the IoU score achieved by a model trained on an anonymized train split and $\text{IoU}_\text{base}$ is the IoU score achieved by the same model trained on the non-anonymized train split.
The iIoU is computed with
\begin{equation}
    \text{iIoU} = \frac{\text{iTP}}{\text{iTP} + \text{FP} + \text{iFN}},
\end{equation}
where FP are false positive pixels, iTP and iFN are instance-weighted true positive and false negative pixels.
Instance weights are determined by computing the contribution of each pixel by the ratio of the class’ average instance size to the size of the respective ground truth instance \cite{cordts2016cityscapes}.

\textbf{Results.} Overall, all models achieve higher IoU scores for the person class and the human category than for the rider class. 
The iIoU scores are lower for all classes; accordingly, large instances are segmented more accurately than small ones.
Naive anonymization methods degrade the IoU score for the person class on average by 0.6\% compared to the baselines, whereas deep learning-based methods only by 0.13\% on average.
For the rider class, the performance drops are more significant.
Naive methods degrade the performance by 2.05\% on average and deep learning-based methods by 0.58\%.
The cityscapes dataset contains an order of magnitude less pixels for the rider class than for the person class.
Hence, it can be inferred that the choice of anonymization method is more important for underrepresented classes or smaller datasets.
Gauss anonymization degrades segmentation with Mask2Former more than with PSPNet.
Crop anonymization, on the other hand, degrades segmentation performance more with PSPNet than with Mask2Former. 
Overall, the deep learning-based methods tend to affect segmentation performance less and are therefore better suited to anonymize datasets in this context.

\begin{table*}[!ht]
    \centering
    \caption{Impacts of anonymization methods on semantic segmentation}
    \label{tab:experiment-seg}
    \resizebox{\textwidth}{!}{\begin{tabular}{llcccccccc}
        \toprule
        Anon. method & Seg. model & $\text{IoU}_{\text{Person}}$ & $\Delta\text{IoU}_{\text{Person}}^{\text{rel}}$ & $\text{iIoU}_{\text{Person}}$ & $\text{IoU}_{\text{Rider}}$ & $\Delta\text{IoU}_{\text{Rider}}^{\text{rel}}$ &  $\text{iIoU}_{\text{Rider}}$ &  $\text{IoU}_{\text{Human}}$ & $\text{iIoU}_{\text{Human}}$ \\
        \midrule
        Baselines \\
        \midrule
        - & \textsc{Mask2Former} & 0.818 & 0.00\% & 0.660 & 0.624 & 0.00\% & 0.492 & 0.830 & 0.694 \\
        - & \textsc{PSPNet} & - & 0.00\% & - & - & 0.00\% & - & - & - \\
        \midrule
        Naive & & & & &\\
        \midrule
        \textsc{Gauss} & \textsc{Mask2Former} &  0.813 &	-0.61\%	&0.653 &0.609&	-2.40\%&	0.479&	0.827&	0.686 \\
        \textsc{Gauss} & \textsc{PSPNet} &  - &	-0.43\%	&- &-&	-0.82\%&	-&	-&	- \\
        \textsc{Crop} & \textsc{Mask2Former} & 0.812	&-0.73\%	&0.639	&0.608	&-2.56\%	&0.490	&0.828	&0.677 \\
        \textsc{Crop} & \textsc{PSPNet} &  - &	-0.99\%	&- &-&	-2.87\% &	-&	-&	- \\
        \textsc{Pixel} & \textsc{Mask2Former} &  0.816	&-0.24\%		&0.643	&0.614	&-1.60\%		&0.495	&0.829	&0.691 \\
        \midrule
        Deep learning-based & & & & & & \\
        \midrule
        \textsc{DeepPrivacy} & \textsc{Mask2Former} & 0.817	& -0.12\% & 0.647 & 0.619 & -0.80\% & 0.497 & 0.830 & 0.686 \\
        \textsc{DeepPrivacy} & \textsc{PSPNet} & - & +0.08\% & - & - & +0.09\% & - & - & - \\
        \textsc{DeepPrivacy2} & \textsc{Mask2Former} &  0.816	&-0.24\%		&0.646	&0.618	&-0.96\%	&0.485	&0.829	&0.686\\
        \textsc{LDFA} (ours) & \textsc{Mask2Former} &  0.816	&-0.24\%	&0.647	&0.620	&-0.64\%	&0.496	&0.831	&0.689\\
        \bottomrule
    \end{tabular}}
\end{table*}

\subsection{Face Detection on Anonymized Images}
While training on synthetic data it is highly important that the artificially created data is close to real-world data.
After applying an anonymization method, persons should not be recognizable compared to the original image, but their faces should still be detectable.
In this experiment we investigate the impacts of face anonymization methods on a recent face detector.

\textbf{Experimental setup.} We run the \textsc{RetinaFace} face detection algorithm on the original Cityscapes train set images.
Detected faces are considered as ground truth. 
In total, 3765 faces are detected in the train set. 
Faces can be categorized into a small (face $< 32^2$ pixels), medium (face $< 32^2$ pixels $\wedge$ face $> 96^2$ pixels) and large (face $\geq 96^2$ pixels) category.
Since the cameras are mounted on a test vehicle, there is always a safety distance between camera and pedestrians. 
As a result, the train set is heavily skewed towards smaller people, and thus smaller faces. (3214 small, 543 medium, and 8 large faces).
We apply each anonymization method to the original image and then repeat the face detection.

\textbf{Evaluation metrics.} We use the mAP as evaluation metric and provide $\text{mAP}_{\text{S}}$, $\text{mAP}_{\text{M}}$, and $\text{mAP}_{\text{L}}$ for the small, medium and large face category. 
In our evaluation, we assume that a high mAP corresponds to the fact, that from an algorithmic viewpoint, a face is still detected as a face after anonymization.
Furthermore, an essential aspect of anonymization is that all persons within an image have to be anonymized. 
Hence, we further evaluate the methods by the number of anonymized faces (NoA).
\begin{table}[ht]
    \centering
    \caption{Impacts of anonymization methods on face detection}
    \label{tab:experiment-face}
    \resizebox{0.48\textwidth}{!}{\begin{tabular}{llllllll}
        \toprule
        Model & mAP & $\text{mAP}_{\text{S}}$ & $\text{mAP}_{\text{M}}$ & $\text{mAP}_{\text{L}}$ & \\
        \midrule
        Naive & & & &\\
        \midrule
        \textsc{Gauss} & 0.3728 & 0.3041 & 0.6991 & 0.4673 & \\
        \textsc{Crop} & 0.0012 & 0.0000 & 0.0034 & 0.0000 &  \\
        \textsc{Pixel} & 0.2153 & 0.1697 & 0.4221 & 0.2792 & \\
        \midrule
        Deep learning-based & & & & & NoA\\
        \midrule
        \textsc{DeepPrivacy} & 0.5661 & 0.5320 & 0.6568 & \textbf{0.7943} & 1878 \\
        \textsc{DeepPrivacy2} & 0.5206 & 0.4738 & 0.6657 & 0.0874 & \textbf{3765} \\
        \textsc{LDFA} (ours) & \textbf{0.6754} & \textbf{0.6652} & \textbf{0.6930} & 0.3168 & \textbf{3765} \\
        \bottomrule
    \end{tabular}}
\end{table}

\begin{figure}[!ht]
    \centering
    \includegraphics[width=\linewidth]{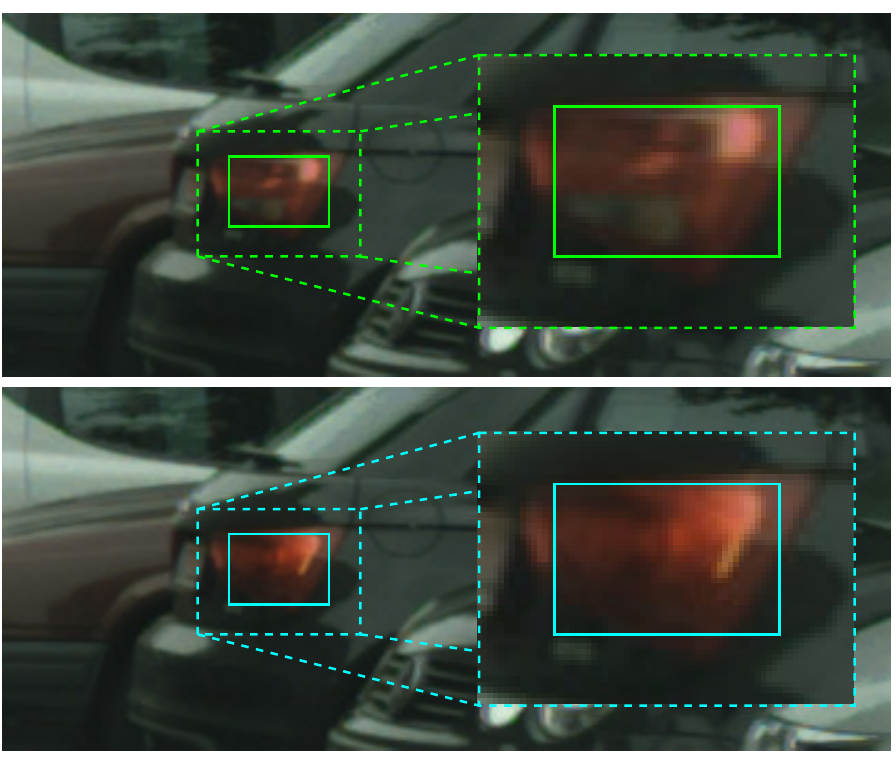}
    \caption{\textbf{An indicator which is falsely detected as face}. The solid line is the detected bounding box and the dotted line is a crop-out for visualization purpose. The Figure shows that our pipeline can accurately generate even non-facial contents without any specialized fine-tuning.}
    \label{fig:false-positives}
\end{figure}

\textbf{Results.} Table \ref{tab:experiment-face} shows the evaluation results of this experiment. 
Naive methods heavily change facial contents without maintaining enough semantic properties for successfully detecting faces after anonymization, resulting in a low mAP.
In contrast, deep learning based methods attempt to generate semantically accurate faces.
The impact of that is repesented by the substantial improvement in mAP in all deep learning based methods compared to naive methods. 
Our method outperforms both \textsc{DeepPrivacy}, \textsc{DeepPrivacy2} and all naive methods in mAP, $\text{mAP}_{\text{S}}$, and $\text{mAP}_{\text{M}}$.
One assumption for this is that LDMs generally perform better for the generation of smaller faces compared to GANs.
Furthermore, our method detects nearly twice as many faces shown by NoA compared to the default settings of \textsc{DeepPrivacy}.
A high NoA may result wrongly detected faces.
However, compared to GAN based methods, an LDM which is trained on a general dataset can infer an alternative version of a false positive.
Whereas a GAN would infer a face within the falsely detected face.
An example of this behaviour is shown in Figure \ref{fig:false-positives}.

\subsection{Embedding Distance as Measure for Anonymization Level}
In this experiment we validate the face recognition ability after applying a face anonymization method.
We use a recent face recognition framework to analyze face embeddings.
These embeddings are high dimensional latent representations of faces generated by a convolutional neural network designed.
Various distance metrics can be used between the face embeddings to determine similar faces. 
In contrast to face recognition, for anonymization, embeddings of a face and  of the corresponding anonymized version should be dissimilar.

\textbf{Experimental setup.} We use the LightFace \cite{serengil2021lightface} framework for detecting and analyzing faces.
As in previous experiments, RetinaFace is used for detection.
Face embeddings are generated using a VGG-Face \cite{parkhideep} model.
Respectively, the generated embeddings are 2622 dimensional vectors.

\textbf{Evaluation metrics.} As a similarity measure, we use L2 normalized euclidean distance (L2), since it is considered as a robust metric in this context \cite{serengil2020lightface}.
The L2 distance is calculated by first l2-normalizing both embedding vectors and then calculating the euclidean distance.
Initially, faces are detected in the original images from the Cityscapes train set. 
We then compare these facial areas with those from the anonymized images and calculate the metrics over the entire dataset.

\textbf{Results.} Overall, the distances are relatively small, which is due to the fact that the majority of faces in this dataset is very small (face $< 32^2$ pixels).
Figure \ref{fig:embedding-dist-hist} shows the distance for all metrics in a histogram. 
\textsc{DeepPrivacy1} and our method have a comparable distribution over the L2 distance.
But there is a big peak coming from images with where $L_2=0$. 
We assume this peak results from faces, which are not detcted by the \textsc{DeepPrivacy} pipeline and therefore not anonymized at all.
The distribtion of \textsc{DeepPrivacy2} is shifted to the right in comparison to our method. 
This may be caused by the poor quality of the in-paintings, where unrealistic faces with mismatched skin colors are created (cf. Figure \ref{fig:matrix}).
The pixelation method yields the highest distances, which makes sense, because the faces should not be recognizable at all.
This means, there is a trade-off between realistic anonymization and making it difficult for CNNs to recognize anonymized faces.

{\tiny
\begin{figure}[htbp]
  \centering
  \includegraphics[width=0.95\linewidth]{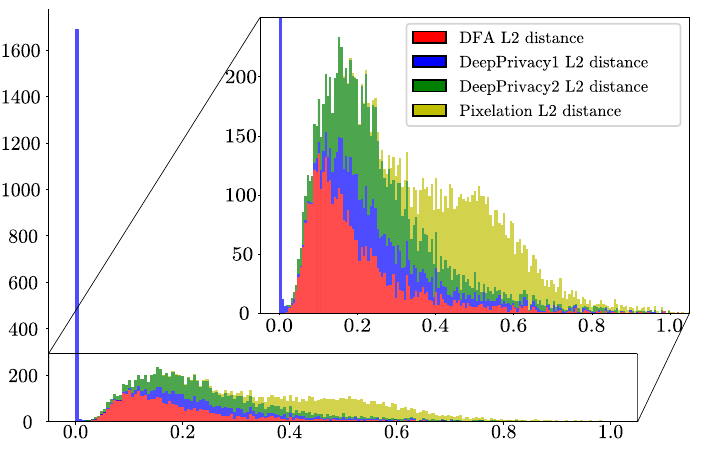}
  \caption{\textbf{Embedding distance as measure for anonymization level.} The histogram shows the L2 distance of each anonymization method over the dataset.}
  \label{fig:embedding-dist-hist}
\end{figure}
}

\subsection{Limitations}
\begin{figure}
\centering
    \includegraphics[width=\linewidth]{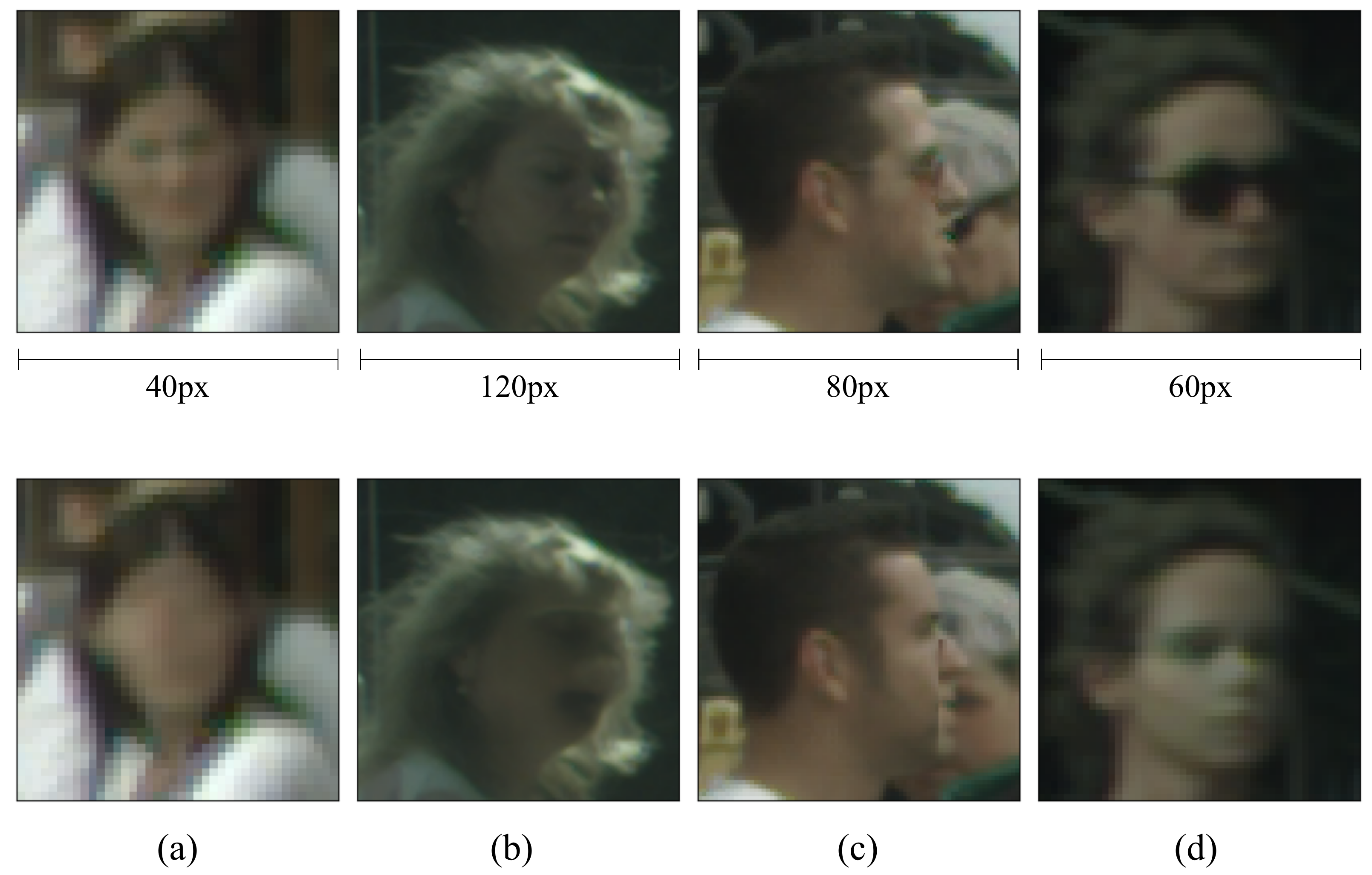}
    \caption{\textbf{Limitations of our pipeline}. The top row shows the input images, the bottom row the corresponding face in-painting generated by our method.}
    \label{fig:limitations}
\end{figure}

As shown in Table \ref{tab:experiment-face}, face detectors achieve higher mAP scores for faces anonymized by our pipeline compared to naive or recent GAN-based anonymization methods. 
Nevertheless, there is still a difference of 0.3246 in the mAP scores compared to the ground truth from non-anonymized faces.
Figure \ref{fig:limitations} shows some limitations of our pipeline that can lead to this discrepancy.
For small faces (area $<$ $32^2$ pixels), the LDM used in our pipeline may generate in-paintings that resemble blurry faces (a).
In rare cases, the used LDM generates severely deformed faces (b).
Accordingly, these faces can no longer be detected after anonymization.
The subfigures (c) and (d) show limitations that do not affect detection, but appear unrealistic.
Overlapping bounding boxes may lead to artifacts (c). 
The removal of sunglasses or glasses may generate in-paintings with mismatching skin colors (d). 

\section{Conclusion}
\label{sec:conclusion}
We presented a two stage anonymization pipeline that integrates state-of-the-art diffusion models into an anonymization task. 
We clearly exhibited that our method is better suited for anonymization, because the recall of the first stage face recognition can be tuned. 
This leads to a higher NoA than comparable methods.
In combination with the general LDM the impact of false positives on the overall anonymization task is negligible. 
To compare anonymization methods, we trained segmentation methods on anonymized data and evaluated them on non-anonymized data. 
Our experiment revealed that our method is better suited to anonymize data for segmentation than naive methods and performed comparably with recent GAN-based methods.
Additionally, we showed that the mAP score for face recognition tasks improves drastically in comparison to other recent GAN-based methods.
The success on face anonymization indicates that this method could be extended to full-body anonymization. 
This should reduce the recognition capability significantly, but the impact on segmentation tasks needs to be investigated.

{\small
\bibliographystyle{ieee_fullname}
\bibliography{11_references}
}

\ifarxiv \clearpage \appendix
\label{sec:appendix}

 \fi

\end{document}


\title{\paperTitle \\ Supplemental Material}
\author{\authorBlock}
\maketitle

\appendix
\label{sec:appendix}


{\small
\bibliographystyle{ieee_fullname}
\bibliography{11_references}
}